\documentclass[runningheads]{llncs}

\usepackage{eccv}

\usepackage{eccvabbrv}

\usepackage{graphicx}
\usepackage{booktabs}
\usepackage[accsupp]{axessibility}  

\usepackage{hyperref}

\usepackage{orcidlink}
\begin{document}

\title{WilLaGS: Latent-Conditional 3D Appearance Fields for Robust Gaussian Splatting In-the-Wild} 

\titlerunning{WilLaGS}

\author{Yuhao Bai\inst{1}\orcidlink{0009-0004-0970-9136} \and
Qianqiu Tan\inst{2}\orcidlink{0009-0007-8641-1810} \and
Lilong Chen\inst{1}\orcidlink{0009-0002-5109-6308} \and
Huanhuan Lv\inst{1}\orcidlink{0009-0003-5296-6501} \and
Lijun Chen\inst{1}\thanks{Corresponding author.}\orcidlink{0009-0003-9062-2304}}

\authorrunning{Y. Bai et al.}

\institute{Nanjing University, China
 \and
Nanjing Agricultural University, China
}
\maketitle
\begin{center}
    \centering
    \captionsetup{type=figure}
    \includegraphics[width=\linewidth]{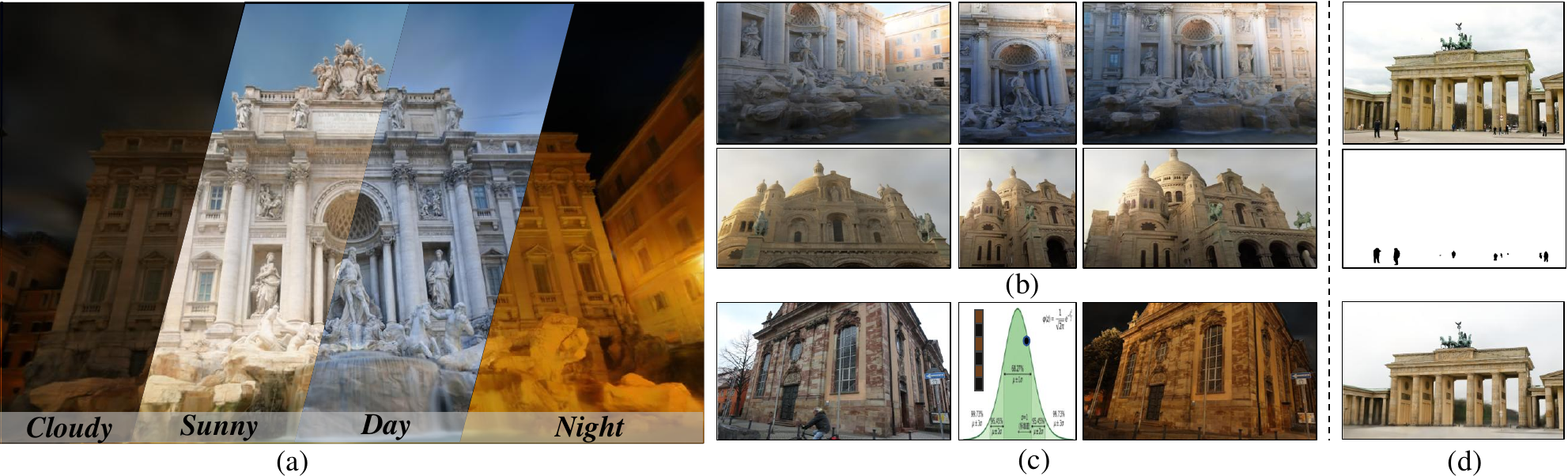}
    \captionof{figure}{\textbf{WilLaGS} enables robust in-the-wild reconstruction and generative appearance modeling, featuring: (a) diverse appearance rendering, (b) strict multi-view consistency, (c) unconditional novel appearance synthesis, and (d) robust transient object removal.}
    \label{abstract}
\end{center}%

\begin{abstract}
3D Gaussian Splatting (3DGS) delivers real-time and high-fidelity rendering but remains challenged by unconstrained in-the-wild scenes, where drastic appearance variations and transient objects violate multi-view consistency. Existing methods are fundamentally limited by independent and discrete embeddings that struggle to capture continuous environmental changes or model spatially-varying local illumination. To address these limitations, we propose \textbf{WilLaGS}, a unified framework for robust 3D scene reconstruction  and generative appearance synthesis under unconstrained settings. 
Specifically, we introduce a generative appearance model where a $\beta$-VAE learns a structured and continuous manifold of global appearance. Conditioned on the latent code, we construct a 3D neural appearance field that generates dynamic Tri-Plane features to encode spatially-varying local illumination effects. Furthermore, to suppress transient artifacts, we present a self-supervised perceptual masking mechanism that leverages a Teacher-Student (EMA) architecture to derive a stable scene consensus, robustly identifying inconsistent regions via perceptual discrepancies. 
Extensive experiments on multiple datasets demonstrate that \textbf{WilLaGS} achieves state-of-the-art performance in reconstruction quality and novel view appearance synthesis, while maintaining real-time rendering efficiency.
\end{abstract}

\section{Introduction}
\label{sec:intro}

Novel view synthesis from image collections is a fundamental challenge in computer vision and graphics, pivotal for immersive applications like VR/AR, robotics, and digital content creation. Recent advances such as implicit Neural Radiance Fields (NeRF) \cite{mildenhall2021nerf,barron2021mipnerf,barron2022mipnerf360,muller2022NGP,barron2023zipnerf,rematas2022urban} and explicit 3D Gaussian Splatting (3DGS) \cite{kerbl20233d,xu2024wildgs,yu2024mipgs,ye2024absgs,zhou2024featuregs,yu2024gsopacity}, have achieved high-quality scene reconstruction and synthesis. In particular, the explicit representation of 3DGS, coupled with an efficient differentiable rasterizer, enables fast convergence and real-time rendering.

Despite these successes, standard 3DGS assume static scenes with consistent illumination and minimal occlusion. However, in-the-wild image collections are inherently unconstrained, featuring diverse lighting, weather, and transient elements like pedestrians or vehicles. These factors present two coupled challenges: (1) severe appearance variations that disrupt photometric consistency across views, and (2) pervasive transient objects that occlude static scene observations. Such inconsistencies violate the multi-view assumptions of 3DGS, leading to severe artifacts such as ghosting, blur, and distorted geometry \cite{sabour2023robustnerf,zhang2024gsw}. 

Existing methods typically treat appearance modeling and transient suppression as decoupled problems. For appearance modeling, most approaches employ per-image appearance embeddings to compensate for lighting and color  inconsistencies \cite{martin2021nerfw,chen2022hanerf,kassab2023refinedfields,li2023nerfms,tancik2022blocknerf,kulhanek2024wildgaussians}. However, a single global embedding is fundamentally ill-suited to the continuous nature of real-world illumination, failing to capture spatially-varying effects such as localized highlights and complex cast shadows. Furthermore, since these discrete embeddings are independently optimized for each training image, they tend to memorize specific appearances rather than learning a generalizable manifold.
For transient handling, current techniques face a central trade-off. Concurrently optimizing auxiliary networks (e.g., U-Net \cite{ronneberger2015unet}) for uncertainty or visibility map estimation often introduces a critical convergence mismatch with the rapid 3DGS optimization, leading to unstable training dynamics \cite{martin2021nerfw,zhang2024gsw,xu2024wildgs,wang2024we}. Conversely, leveraging external pre-trained models (e.g., LSeg \cite{li2022lseg} or SEEM \cite{zou2023segment}) for explicit masking sacrifices end-to-end adaptability and struggles to capture dataset-specific transients \cite{wang2025look}.

To address these limitations, we propose \textbf{WilLaGS}, a novel and unified framework for robust in-the-wild 3D reconstruction.
Our key insight is to formulate appearance modeling not as a deterministic per-image embedding, but as a probabilistic structure (e.g., a Gaussian prior) to learn a shared and generative manifold of global appearance.
Specifically, we employ a $\beta$-Variational Autoencoder ($\beta$-VAE) \cite{kingma2013vae,higgins2017beta} to map complex environmental appearance variations into a structured and continuous latent space. This representation effectively distills diverse environmental conditions into a compact generative prior, avoiding the overfitting pitfalls inherent to unconstrained per-image embeddings.
To capture spatially-varying illumination (e.g., shadows), we introduce a latent-conditioned 3D neural field. Conditioned on the global latent code, this field generates dynamic Tri-Plane features to spatially modulate the color of Gaussian primitives, effectively bridging global style with local illumination. 
Finally, for transient suppression, we propose a self-supervised masking mechanism based on a Teacher-Student architecture. By leveraging temporal consistency via Exponential Moving Average (EMA), our teacher model aggregates a stable scene consensus, allowing us to robustly identify and filter transient regions via perceptual discrepancies. Derived directly from training dynamics, this stable consensus effectively bypasses the instability of learnable uncertainty modules and the generalization limits of pre-trained segmentation models.

Our main contributions are summarized as follows:
\begin{itemize}
\item We propose \textbf{WilLaGS}, a novel and unified framework for robust scene reconstruction under unconstrained conditions, effectively handling extreme appearance variations and transient objects.
\item We introduce a $\beta$-VAE to impose a probabilistic structure and learn a shared, generative manifold of global appearance, combined with a latent-conditioned 3D neural field to capture spatially-varying local illumination.
\item We present a self-supervised perceptual masking mechanism based on a Teacher-Student architecture, using temporal averaging to generate a stable static scene consensus and perceptual feature discrepancies to identify transient regions.
\item Extensive experiments on multiple in-the-wild scenes demonstrate the superiority of our approach in scene reconstruction and novel view synthesis.
\end{itemize}

\section{Related Work}
\subsection{Appearance Variations Modeling in Neural Rendering.}  Reconstructing 3D scenes from unconstrained photo collections remains challenging due to severe appearance variations from diverse illumination, weather, and camera settings \cite{li2020crowdsampling}. Pioneering work like NeRF-W \cite{martin2021nerfw} introduced learnable per-image appearance embeddings to compensate for photometric variations. Subsequent works, including Ha-NeRF \cite{chen2022hanerf} and CR-NeRF \cite{yang2023cross} employed CNNs \cite{krizhevsky2012cnn} or Transformers \cite{vaswani2017attention} to encode appearance from images, albeit with high computational cost and slow convergence.
More recent efforts extend 3DGS to unconstrained settings, leveraging its real-time rendering efficiency. Scaffold-GS \cite{lu2024scaffold} and SWAG \cite{dahmani2024swag} embed image-conditioned features into each Gaussian primitives to model view-dependent illumination, while WE-GS \cite{wang2024we} employed a spatial attention for adaptive embeddings prediction. GS-W \cite{zhang2024gsw} and Wild-GS \cite{xu2024wildgs} decomposed illumination hierarchically to separate global lighting and local reflectance. Further, WildGaussians \cite{kulhanek2024wildgaussians} and Splatfacto-W \cite{xu2024splatfacto} extended per-Gaussian attributes with additional appearance MLPs to simulate affine color transformations. 
Although effective, these methods rely on discrete per-image or per-Gaussian embeddings, which cannot capture spatially varying illumination or generate novel continuous appearance variations. In contrast, our WilLaGS organizes in-the-wild appearance diversity into a generative continuous latent space learned via a $\beta$-VAE, and conditions a 3D neural appearance field to model spatially-aware and physically consistent illumination.

\subsection{Transient Object Suppression and Robust Reconstruction.} Another hurdle in in-the-wild reconstruction is the presence of transient objects like pedestrians and vehicles, which violate static scene assumptions \cite{zhao2022factorized}. Existing approaches can be broadly divided into two categories.

\textit {Joint optimization with auxiliary networks}. Methods such as NeRF-W \cite{martin2021nerfw}, Ha-NeRF \cite{chen2022hanerf}, IE-NeRF \cite{wang2025ie}, GS-W \cite{zhang2024gsw}, and Wild-GS \cite{xu2024wildgs} train auxiliary modules (e.g., U-Net \cite{ronneberger2015unet} or MLP \cite{rosenblatt1958perceptron}) to predict per-image uncertainty or visibility maps. To mitigate the instability of pixel-wise signals under appearance changes, NeRF On-the-go \cite{ren2024nerfonthego}, SLS \cite{sabour2025spotlesssplats}, and WildGaussian \cite{kulhanek2024wildgaussians} leverage semantic features (e.g., DINO \cite{caron2021diino,oquab2024dinov2}) for more robust uncertainty estimation.
However, these methods suffer from convergence mismatch, as the slow optimization of the auxiliary network bottlenecks the rapid convergence of 3DGS. 

\textit{Explicit masking using pre-trained segmentation models}. Works such as NeRF-HuGS  \cite{chen2024nerfhugs} and Look at the Sky \cite{wang2025look} employ pre-trained models (e.g., SAM \cite{zou2023segment}) to explicitly segment and remove transient regions. While effective, these approaches lack end-to-end adaptability and often inherit dataset-specific biases.

 In contrast, our WilLaGS introduces a self-supervised mechanism based on perceptual consistency within a teacher-student framework. This automatically identifies inconsistent regions and generates robust masks tailored to the scene learned consensus without external priors.

\section{Method}

\begin{figure*}[t]
  \centerline{\includegraphics[width=\linewidth]{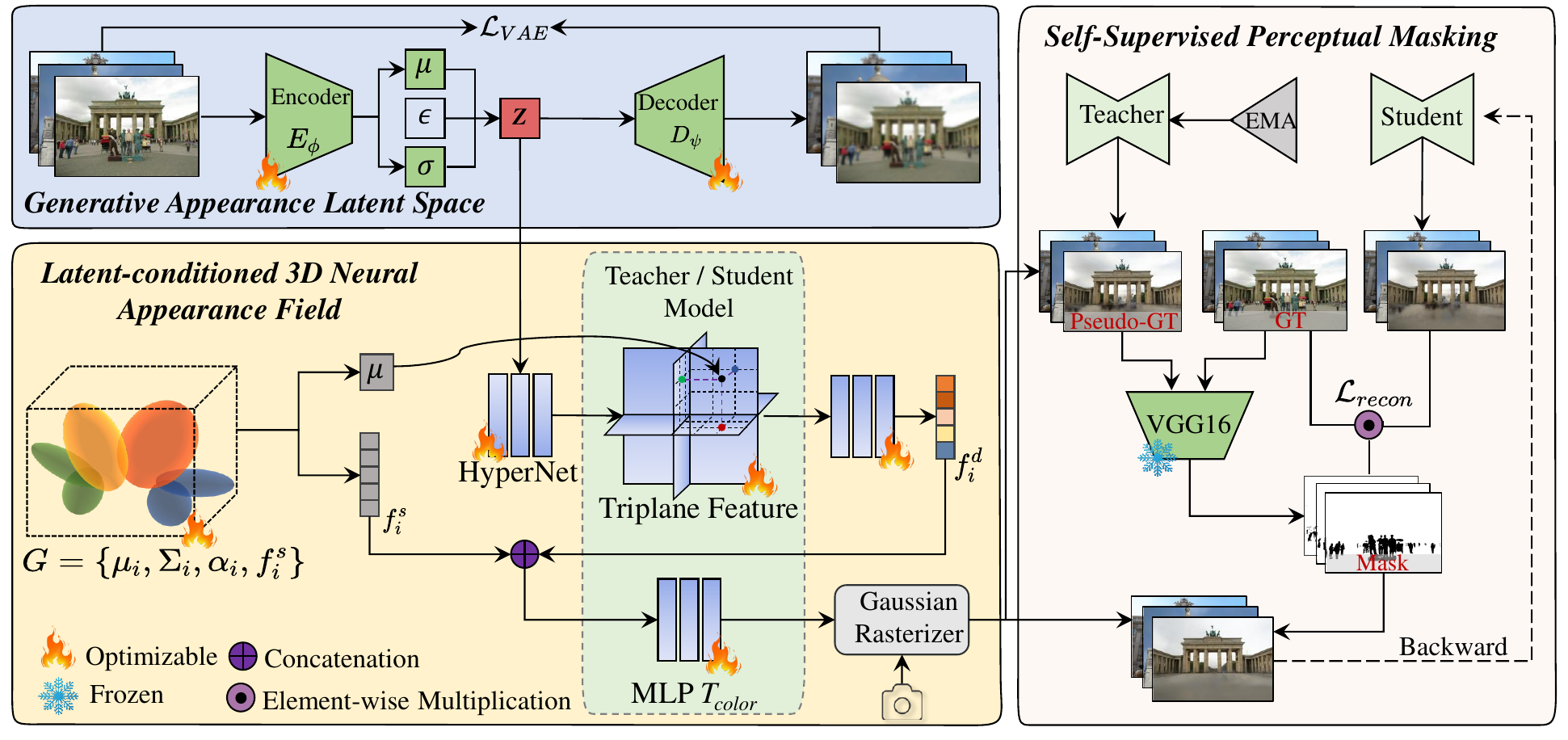}}
  \caption{Overview of our \textbf{WilLaGS} framework. Given unconstrained multi-view images, WilLaGS first learns a generative latent space via a $\beta$-VAE to capture the global appearance manifold. Conditioned on the latent code $z$, a 3D neural appearance field based on a Tri-Plane representation is constructed to modulate spatially-varying local illumination. Finally, a self-supervised perceptual masking module leveraging a Teacher-Student architecture is introduced to robustly suppress transient objects.}
\label{flowchart}
\end{figure*}

We propose \textbf{WilLaGS}, a novel and self-supervised framework for robust 3DGS reconstruction from unconstrained image collections (Fig. \ref{flowchart}). Our method jointly addresses appearance variation and transient objects through three synergistic components. First, we introduce a generative appearance latent space learned via a $\beta$-VAE, capturing the structured and continuous manifold of global appearance (Sec. \ref{sec:vae}). Second, a latent-conditioned 3D neural appearance field is developed to produce spatially-aware features and local illumination (Sec. \ref{sec:field}). Finally, we propose a self-supervised teacher–student masking mechanism that exploits perceptual discrepancies between the ground truth and the EMA teacher-generated scene consensus to identify transient regions while remaining robust to severe appearance variations (Sec. \ref{sec:mask}).

\subsection{Preliminaries: 3D Gaussian Splatting (3DGS)}
3DGS \cite{kerbl20233d} represents a scene as a set of anisotropic 3D Gaussian primitives, each defined by its center position $\mu_i\in \mathbb{R}^3$, a positive semi-definite covariance matrix $\Sigma_i\in\mathbb{R}^{3\times3}$, an opacity $\alpha_i\in\mathbb{R}$, and a view-dependent color parameterized using spherical harmonics (SH). During rendering, these 3D Gaussians are projected onto the image plane, sorted by depth, and accumulated through a differentiable tile-based rasterizer using alpha compositing. Formally, the color $C$ for a pixel is computed as:
\begin{equation}
\begin{aligned}
  C &= \sum_{i=1}^N c_i\, \alpha_i^{\prime} \prod_{j=1}^{i-1} (1 - \alpha_j^{\prime}), \\
  \alpha_i^{\prime} &= \alpha_i \cdot \exp\!\left(-\tfrac{1}{2}(x - \mu_{i}^{\prime})^{T}(\Sigma_{i}^{\prime})^{-1}(x - \mu_{i}^{\prime})\right)
\end{aligned}
\label{eq:gs}
\end{equation}
where, $c_i$ denotes the view-dependent color of the $i$-th Gaussian, $\mu_i^{\prime}$ and $\Sigma_i^{\prime}$ are the mean and covariance of the Gaussian projected onto the 2D image plane.
\subsection{Generative Appearance Latent Space}
\label{sec:vae}
To overcome the limitations of discrete per-image embeddings, we model appearance variations using a continuous generative latent space learned via a $\beta$-Variational Autoencoder ($\beta$-VAE) \cite{higgins2017beta}. This latent space captures the underlying manifold of global appearance factors (e.g., illumination, weather) and provides a continuous representation that generalizes beyond the training set.

The $\beta$-VAE consists of an encoder $E_{\phi}$ that maps an input image $I_j$ to a latent distribution $q_{\phi}(z|I_j) = \mathcal{N}(\mu_j, \sigma_j^2)$, and a decoder $D_{\psi}$ that reconstructs the image $\hat{I}_j$ from a sampled latent code $z_j$. The $\beta$-VAE is optimized with a weighted ELBO objective \cite{blei2017variational}:
\begin{equation}
\mathcal{L}_{\text{VAE}} = \mathcal{L}_{\text{recon}}(I_j, \hat{I}_j) + \beta \cdot D_{\text{KL}}\big(q_{\phi}(z|I_j) \,\|\, p(z)\big)
\end{equation}
where $\mathcal{L}_{\text{recon}}$ enforces pixel-wise reconstruction fidelity (e.g., L1 loss),  and the $D_{KL}$ term regularizes the latent posterior $q_{\phi}(z|I_j)$ towards a standard Gaussian prior $p(z) = \mathcal{N}(0, I)$. 
The hyperparameter $\beta$ controls the trade-off between reconstruction fidelity and the structure of the latent space. Increasing $\beta$ imposes a stronger information bottleneck, encouraging the learned latent distribution to be compact and statistically independent. Unlike standard VAEs ($\beta=1$), a higher $\beta$ suppresses the encoding of redundant correlations and high-frequency noise. In our work, we set $\beta > 1$ to prioritize learning a structured and smooth latent manifold over pixel-perfect reconstruction. This regularization strategy prevents the model from memorizing image-specific artifacts (overfitting) and instead compels it to capture essential global variations (e.g., illumination shifts) in a continuous manner, yielding a representation that is robust and generative. 


\subsection{Latent-conditioned 3D Appearance Field}
\label{sec:field}
A single global latent code is insufficient to capture spatially varying illumination effects, such as localized highlights or cast shadows. To model such effects, we introduce a latent-conditioned 3D neural appearance field that maps the global appearance latent code $z_j$ to a spatially-aware representation via a hypernetwork, enabling the localized expression of global style variations.

Specifically, we employ a hypernetwork $H$ that takes the $z_j \in \mathbb{R}^d$ as input and outputs the parameters $\Theta_{field}(z_j)$ for a lightweight 3D neural field. To balance efficiency and spatial expressiveness, we adopt the Tri-Plane representation as appearance field, where $\Theta_{field}$ consists of three orthogonal feature planes.
\begin{equation}
\{P_{XY}, P_{XZ}, P_{YZ}\} = \text{Hypernetwork}(z_j)
\label{eq:hypernetwork}
\end{equation}
This model allows the global latent code $z_j$ to dynamically modulate the entire 3D appearance field.
Each Gaussian can then acquire a unique appearance feature $f_i^d$ by querying this 3D field using its spatial position. For a Gaussian primitive at position $\mu_i$, we first normalize its coordinates to $\mu_{i}^{\prime} \in [-1, 1]^3$ based on the scene bounds. The normalized point is then projected onto the three orthogonal planes, and plane-wise appearance features are sampled at the corresponding 2D coordinates using bilinear interpolation.

\begin{equation}
\begin{aligned}
f_{XY} &= \text{Sample}(P_{XY}, (x_i', y_i')) \\
f_{XZ} &= \text{Sample}(P_{XZ}, (x_i', z_i')) \\
f_{YZ} &= \text{Sample}(P_{YZ}, (y_i', z_i'))
\end{aligned}
\label{eq:sample_planes}
\end{equation}
The sampled vectors are concatenated and passed through a small MLP $T_{\theta}$ to produce the dynamic feature $f_i^d $ for each Gaussian:
\begin{equation}
f_i^d = T_{\theta} (f_{XY}, f_{XZ}, f_{YZ}) 
\end{equation}
This 3D coordinate-based query mechanism allows each Gaussian to acquire spatially varying appearance information modulated by the global latent code, effectively capturing non-uniform illumination across the scene.

To preserve the intrinsic appearance properties of the scene, we adopt an appearance disentanglement strategy inspired by GS-W \cite{zhang2024gsw}. Each Gaussian also maintains a learnable intrinsic feature $f_i^s$ that encodes material-related attributes such as albedo and reflectance, which remain invariant across illumination changes. The final view-dependent color $C_i$ is computed by a lightweight fusion decoder $T_{color}$ that integrates the dynamic feature $f_i^d$, intrinsic feature $f_i^s$ and the view direction $d$: 
\begin{equation}
C = T_{\text{color}}(f_i^d, f_i^s, d)
\end{equation}
This design enables global appearance latent codes to modulate local illumination variations without altering intrinsic material properties, thereby achieving a more disentangled and controllable representation of scene appearance.
\subsection{Self-Supervised Transient Masking}
\label{sec:mask}
Transient objects (e.g., moving pedestrians or vehicles) violate static-scene assumptions and cause reconstruction artifacts. To handle them without external models or extra supervision, we propose a novel self-supervised masking mechanism integrated within a teacher-student framework guided by perceptual consistency. This design allows the model to leverage its own evolving temporal consistency to automatically identify and suppress transient regions.
\begin{figure}[t]
  \centerline{\includegraphics[width=\linewidth]{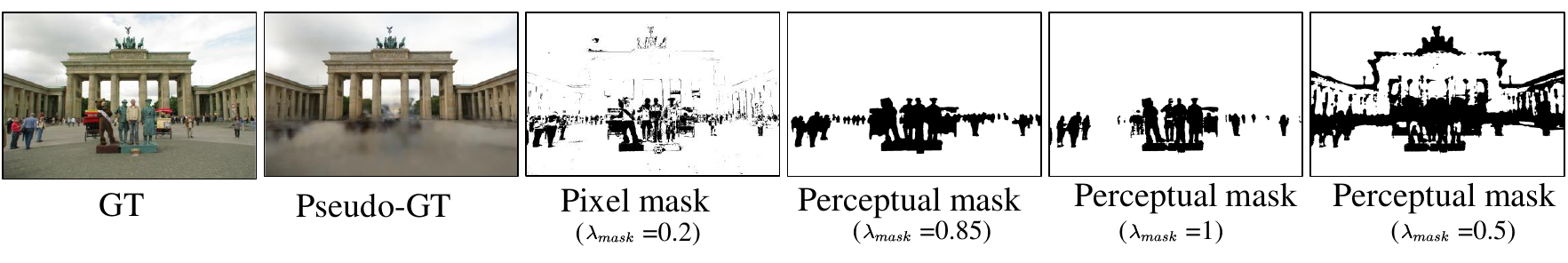}}
  \caption{Comparison between pixel-level and perceptual mask.}
  \label{mask}
  \vspace{-10pt} 
\end{figure}

\textbf{Teacher-Student Architecture}. Our framework comprises a student model and a teacher model. Both share an identical architecture, including our 3D appearance field and color decoder ($T_{color}$). The student model is updated directly via the optimization loss in each iteration, while the teacher parameters $W_T$ are updated as an EMA \cite{hunter1986ema} of the student weights $W_S$:
\begin{equation}
    W_T \leftarrow \tau W_T + (1 - \tau) W_S
\end{equation}
where $\tau$ is a decay factor. The teacher model $W_T$, updated more slowly, aggregates the student knowledge over time and forms a temporally smoothed consensus of the static scene structure and average appearance. The EMA process effectively filters out high-frequency temporal variations caused by transient elements or rendering noise. During training iteration, this teacher model renders a pseudo-ground truth reference image $\hat{I}_{pseudo}$ for the current view.

\textbf{Perceptual Difference Mask Generation}. Given that pixel-level differences are highly sensitive to appearance details (Fig.~\ref {mask}), we compute the discrepancy between ground-truth image $I_{gt}$ with the teacher reference $\hat{I}_{pseudo}$ in a perceptual feature space. Specifically, both images are fed into a frozen pre-trained VGG \cite{simonyan2015vgg} network, and the L1 difference of feature maps [${\phi_l(I_{gt}), \phi_l(\hat{I}_{pseudo}) }$] is computed across multiple layers $l$. The absolute differences are averaged across channels and summed over layers to yield a perceptual difference map $D_{perceptual}$:
\begin{equation}
    D_{perceptual}(p) = \sum_{l} || \phi_l(I_{gt})(p) - \phi_l(\hat{I}_{pseudo})(p) ||_1
\end{equation}
where the layer set $\phi_l = \{\texttt{relu1\_2}, \texttt{relu2\_2}, \texttt{relu3\_3}\}$.
Subsequently, a binary static mask $M_{static}$ is generated by thresholding the difference map:
\begin{equation}
    M_{static}(p) = (D_{perceptual}(p) < \lambda_{mask})
\end{equation}
where $\lambda_{mask}$ is a predefined threshold. Pixels with low perceptual differences are considered part of the consistent static background, while high-difference regions (presumed transient) are excluded from supervision.
This perceptual masking strategy effectively ignores illumination shifts and focuses on identifying structural and content-level inconsistencies introduced by transient elements.

\begin{figure}[t]
  \centerline{\includegraphics[width=\linewidth]{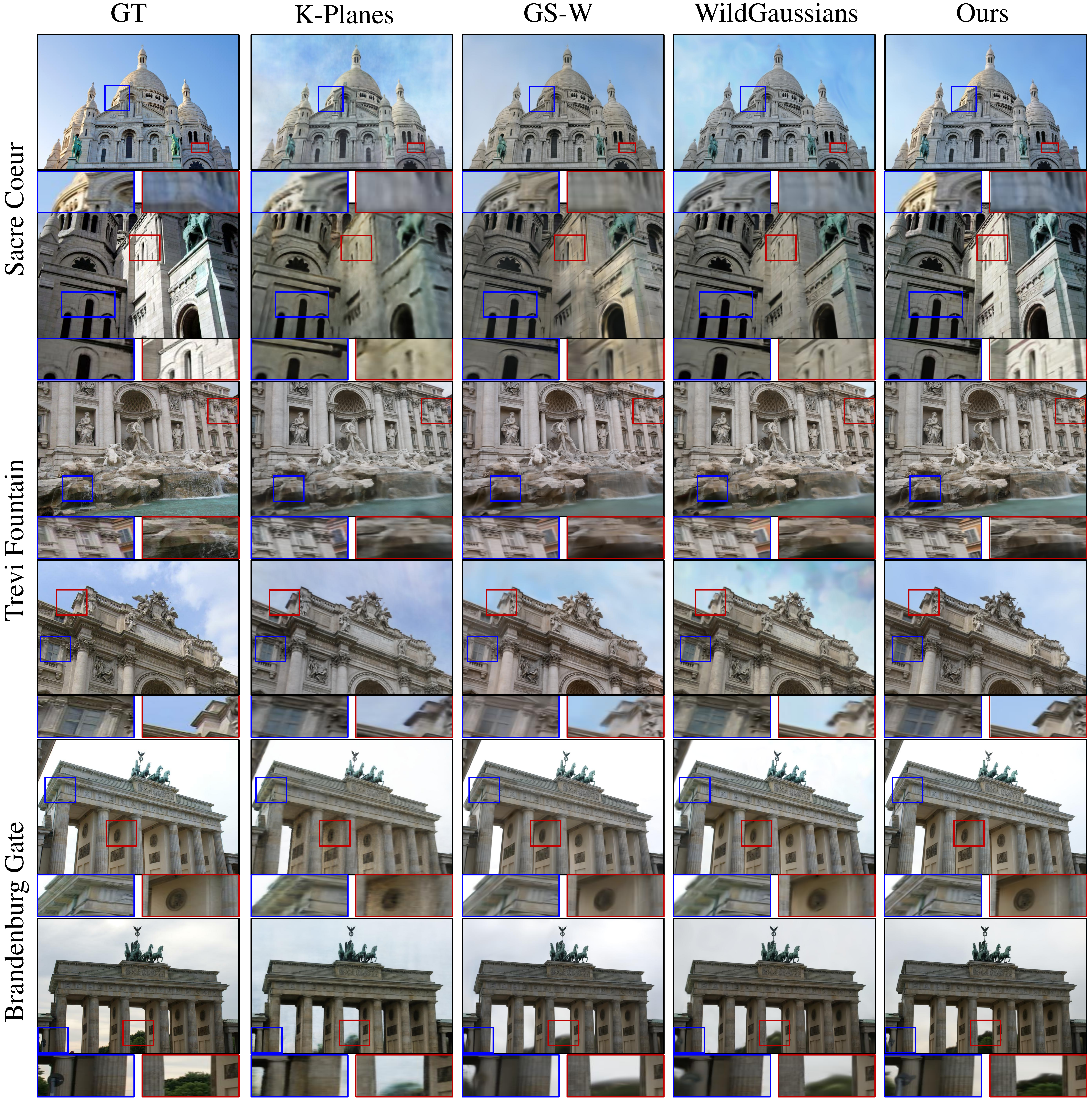}}
  \caption{\textbf{Qualitative comparison on PT Dataset}. WilLaGS recovers finer geometric details (e.g., crisp \textit{Brandenburg Gate} pillars and \textit{Trevi Fountain} facade carvings) and synthesizes spatially-varying local illumination (e.g., realistic localized shadows in \textit{Sacre Coeur} alcoves) while effectively suppressing ghosting artifacts.}
  \label{fig:qual_pt}
\end{figure}

\subsection{Training Objective and Optimization}
Following 3DGS, we use a combination of the L1 loss and the D-SSIM loss computed between the student-rendered image $I_{render}$ and the ground truth image $I_{gt}$. To handle transient objects, the reconstruction loss $\mathcal{L}_{recon}$ is weighted by the dynamically generated static mask $M_{static}$: 
\begin{align}
\mathcal{L}_{\text{recon}} &= 
\lambda_{L1} \, \mathcal{L}_1(I_{\text{render}} \odot M_{\text{static}}, I_{\text{gt}} \odot M_{\text{static}}) \nonumber\\
&\quad + \lambda_{\text{DSSIM}} \, \mathcal{L}_{\text{DSSIM}}(I_{\text{render}} \odot M_{\text{static}}, I_{\text{gt}} \odot M_{\text{static}})
\end{align}
where $\odot$ denotes element-wise multiplication. 
In addition, we incorporate the VAE loss $\mathcal{L}_{VAE}$ to jointly optimize the appearance latent space:
\begin{equation}
    \mathcal{L}_{total} = \mathcal{L}_{recon} + \lambda_{VAE} \mathcal{L}_{VAE}
\end{equation}
 In our study, $\lambda_{L1}$, $\lambda_{DSSIM}$ and $\lambda_{VAE}$ are set to  0.8, 0.2, and 0.01.

\begin{figure}[t]
  \centerline{\includegraphics[width=\linewidth]{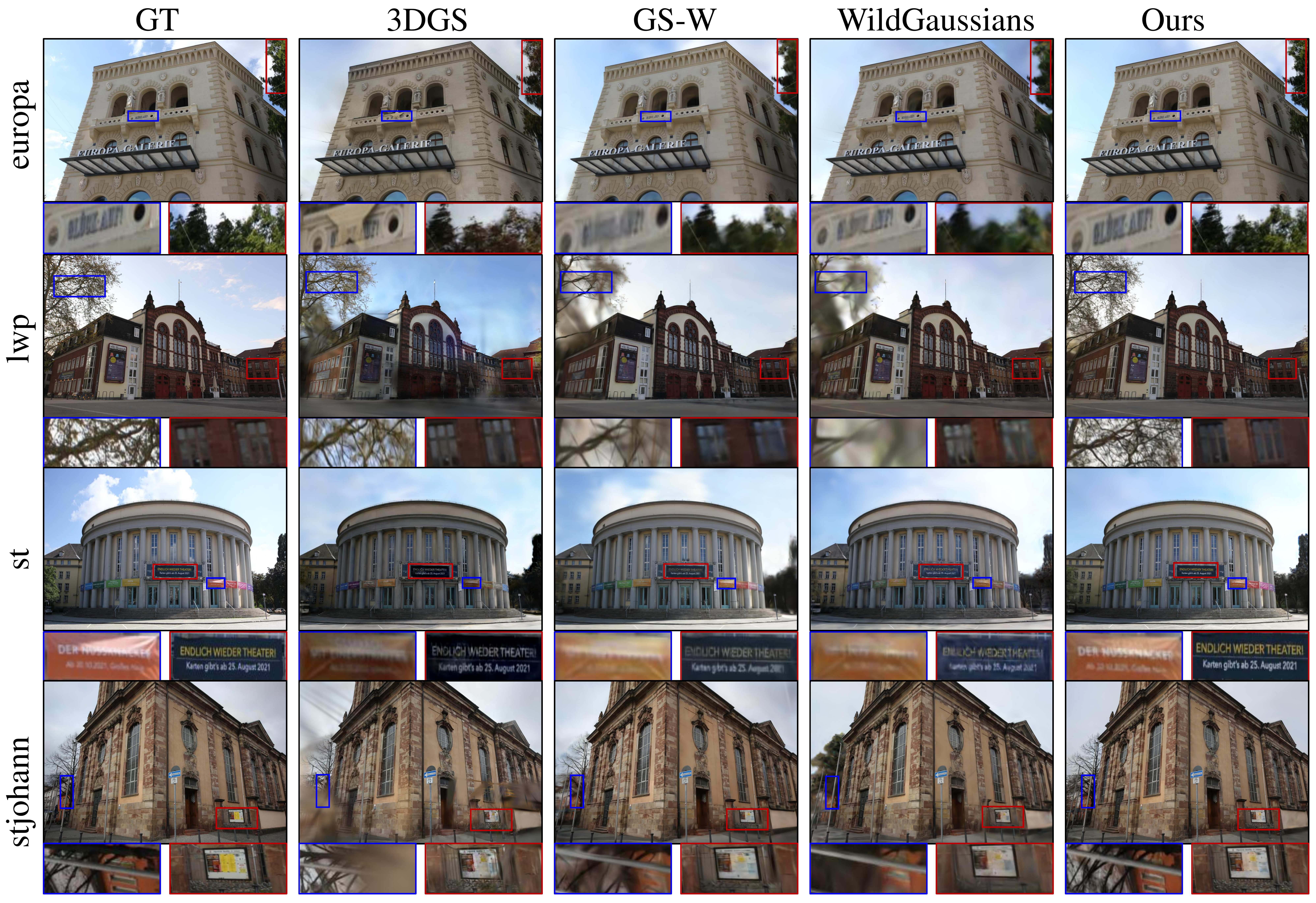}}
  \caption{\textbf{Qualitative comparisons on NeRF-OSR dataset}. WilLaGS excels at preserving challenging high-frequency structures that prior methods typically over-smooth. Notably, our method successfully reconstructs legible text on banners (\textit{st}) and intricate tree branches (\textit{lwp}) while accurately modeling complex outdoor lighting conditions.}
  \label{fig:qual_osr}
\end{figure}

\section{Experiments}
\subsection{Experimental Setup}
\textbf{Datasets}. We evaluate our method on two challenging in-the-wild datasets: Photo Tourism (PT) \cite{snavely2006photo} and NeRF-OSR \cite{rudnev2022nerf}. PT captures large-scale landmarks (e.g., \textit{Brandenburg Gate}) with diverse appearance and heavy transient occlusions. NeRF-OSR features four outdoor scenes (\textit{europa}, \textit{lwp}, \textit{st}, \textit{stjohann}) exhibiting significant illumination and weather variations.

\textbf{Baselines and Evaluation Metrics}. We compare our method against state-of-the-art in-the-wild baselines, including NeRF-based methods (NeRF-W \cite{martin2021nerfw}, Ha-NeRF \cite{chen2022hanerf}, CR-NeRF \cite{yang2023cross}, and K-Planes \cite{fridovich2023klane}) and 3DGS-based methods (SWAG \cite{dahmani2024swag}, GS-W \cite{zhang2024gsw}, Splatfacto-W~\cite{xu2024splatfacto},DeSplat~\cite{wang2025desplat},AsymGS~\cite{li2026asym} and WildGaussians \cite{kulhanek2024wildgaussians}). To ensure strict evaluation fairness, all baselines were evaluated under standardized configurations (e.g., original resolutions, test-time optimization). For novel view synthesis quality, we use standard metrics including PSNR, SSIM \cite{wang2004ssim}, and LPIPS \cite{zhang2018lpips}. Following standard protocols in NeRF-W \cite{martin2021nerfw}, we also optimize the appearance latent code on the left half of each test image and evaluate reconstruction quality on the unseen right half. Additionally, we report training time and rendering FPS to assess computational efficiency.


\textbf{Implementation Details.} We implement our WilLaGS framework using Pytorch and train the proposed networks with Adam optimizer \cite{adam2014adam}.
The $\beta$-VAE encodes images into a latent space of dimension 64, with the KL-divergence weight $\beta$ set to 2.  Both the Tri-Plane Feature Generator and the fusion decoder $T_{color}$ are implemented as lightweight 3-layer MLPs with ReLU activations. The generator maps the latent code $z \in \mathbb{R}^{64}$ to three orthogonal feature planes, each with a spatial resolution of $64 \times 64$ and 32 channels.
For the Gaussian primitives, both the learnable intrinsic feature $f_i^s$ and the queried dynamic feature $f_i^d$ are 32-dimensional. For the perceptual masking, the mask threshold $\lambda_{mask}$ is set to 0.85. Training was conducted on a single Nvidia RTX 3090 GPU for 30k steps. Comprehensive sensitivity analyses justifying the selection of hyperparameters (e.g., $\beta$, $\lambda_{mask}$) and more details are provided in the supplementary material.

\subsection{Comparison Results}

\textbf{Quantitative Results.} Tables \ref{tab:result} and \ref{tab:result_osr} summarize the quantitative comparisons on the PT and NeRF-OSR datasets, respectively. WilLaGS achieves the best or competitive performance in reconstruction fidelity across diverse in-the-wild scenarios. For instance, on the \textit{Trevi Fountain} scene, our method achieves 24.27 dB in PSNR, representing a solid +0.86 dB improvement over the strongest concurrent competitor, AsymGS \cite{li2026asym}. Furthermore, WilLaGS consistently delivers the best or highly competitive SSIM and LPIPS scores across all scenes, demonstrating its superiority in preserving structural integrity under complex environmental shifts. Regarding computational overhead (Table \ref{tab:result}), WilLaGS achieves exceptional quality without bottlenecking training dynamics. Training the full pipeline requires only 0.9 GPU hours per scene while maintaining real-time rendering (58 FPS), making it significantly more efficient than recent baselines like WildGaussians (7.8 hrs) and AsymGS (3.2 hrs).

\begin{table}[t]
\centering
\caption{Quantitative comparison on the PT dataset. 
Best and second-best results are \textbf{bolded} and \underline{underlined}. Our method shows overall superior performance over baseline methods. Results marked with \textsuperscript{*} are directly taken from the original paper.}
\resizebox{\textwidth}{!}{ 
\begin{tabular}{@{} l @{\hspace{6pt}} c@{\hspace{3pt}}c@{\hspace{3pt}}c @{\hspace{6pt}} c@{\hspace{3pt}}c@{\hspace{3pt}}c @{\hspace{6pt}} c@{\hspace{3pt}}c@{\hspace{3pt}}c @{\hspace{6pt}} c @{}}
\hline
 & \multicolumn{3}{c}{Sacre Coeur} & \multicolumn{3}{c}{Trevi Fountain} & \multicolumn{3}{c}{Brandenburg Gate} & GPU hrs./ \\ 
\cline{2-10}
 & PSNR$\uparrow$ & SSIM$\uparrow$ & LPIPS$\downarrow$ & PSNR$\uparrow$ & SSIM$\uparrow$ & LPIPS$\downarrow$ & PSNR$\uparrow$ & SSIM$\uparrow$ & LPIPS$\downarrow$ &  FPS \\
\hline
NeRF-W        & 19.20 & 0.808 & 0.192 & 18.97 & 0.698 & 0.265 & 24.17 & 0.891 & 0.167 & 42/$<$1 \\
Ha-NeRF       & 20.02 & 0.801 & 0.171 & 20.18 & 0.691 & 0.223 & 24.04 & 0.887 & 0.139 & 35/$<$1 \\
CR-NeRF       & 22.07 & 0.823 & \underline{0.152} & 21.48 & 0.712 & \underline{0.207} & 26.53 & 0.900 & \textbf{0.106} & 31/$<$1 \\
K-Planes      & 20.61 & 0.768 & 0.352 & 22.69 & 0.726 & 0.351 & 23.23 & 0.865 & 0.282 & 0.3/$<$1 \\
3DGS          & 17.49 & 0.828 & 0.217 & 17.35 & 0.674 & 0.300 & 19.80 & 0.883 & 0.179 & 0.4/105 \\
SWAG\textsuperscript{*}          & 21.16 & 0.860 & 0.185 & 23.10 & \underline{0.815} & 0.208 & 26.33 & \underline{0.929} & 0.139 & 0.8/15 \\
GS-W          & 19.78 & 0.834 & 0.261 & 20.11 & 0.763 & 0.237 & 23.86 & 0.891 & 0.183 & 2.0/50 \\
Splatfacto-W      & 22.52 & 0.867 & 0.169 & 22.64 & 0.762 & 0.229 & 26.75 & 0.926 & 0.143 & 1.1/40 \\
AsymGS         & \underline{23.29} & 0.862 & 0.169 & \underline{23.61} & 0.763 & 0.243 & \underline{28.19} & 0.924 & 0.146 & 3.2/47 \\
DeSplat      & 20.14 & \underline{0.868} & 0.178 & 23.31 & 0.775 & 0.226 & 25.04 & 0.920 & 0.142 & 2.4/53 \\
WildGaussians & 22.61 & 0.856 & 0.173 & 23.47 & 0.768 & 0.228 & 27.46 & 0.927 & 0.141 & 7.8/73 \\ 
\hline
\textbf{Ours(WilLaGS)} & \textbf{23.80} & \textbf{0.889} & \textbf{0.150} & \textbf{24.27} & \textbf{0.822} & \textbf{0.168} & \textbf{28.45} & \textbf{0.937} & \underline{0.134} & 0.9/58 \\
\hline
\end{tabular}
}
\label{tab:result}
\end{table}

\begin{table}[t] 
\centering
\caption{Quantitative results on the NeRF-OSR dataset. Our WilLaGS outperforms the previous methods across all scenes on PSNR, SSIM, and LPIPS.}
\label{tab:result_osr}
\resizebox{\textwidth}{!}{
\begin{tabular}{@{} l @{\hspace{3pt}} c@{\hspace{3pt}}c@{\hspace{3pt}}c @{\hspace{3pt}} c@{\hspace{3pt}}c@{\hspace{3pt}}c @{\hspace{3pt}} c@{\hspace{3pt}}c@{\hspace{3pt}}c @{\hspace{3pt}} c@{\hspace{3pt}}c@{\hspace{3pt}}c @{}}
\toprule
 & \multicolumn{3}{c}{europa}
 & \multicolumn{3}{c}{lwp}
 & \multicolumn{3}{c}{st}
 & \multicolumn{3}{c}{stjohann} \\
\cmidrule(lr){2-4} 
\cmidrule(lr){5-7} 
\cmidrule(lr){8-10} 
\cmidrule(lr){11-13}
 & PSNR$\uparrow$ & SSIM$\uparrow$ & LPIPS$\downarrow$
 & PSNR$\uparrow$ & SSIM$\uparrow$ & LPIPS$\downarrow$
 & PSNR$\uparrow$ & SSIM$\uparrow$ & LPIPS$\downarrow$
 & PSNR$\uparrow$ & SSIM$\uparrow$ & LPIPS$\downarrow$ \\
\midrule
NeRF-W
& 20.00 & 0.699 & 0.347 
& 19.61 & 0.616 & 0.445 
& 20.31 & 0.607 & 0.438 
& 21.23 & 0.667 & 0.426 \\

Ha-NeRF
& 17.79 & 0.632 & 0.421 
& 20.03 & 0.685 & 0.365 
& 17.30 & 0.538 & 0.483 
& 19.93 & 0.686 & 0.331 \\

CR-NeRF
& 19.92 & 0.696 & \underline{0.310} 
& 20.61 & 0.736 & 0.358 
& 20.47 & 0.661 & 0.390 
& 21.27 & 0.815 & 0.275 \\

3DGS
& 17.71 & 0.728 & 0.311 
& 15.41 & 0.686 & 0.337 
& 16.09 & 0.667 & \underline{0.356} 
& 16.17 & 0.744 & 0.289 \\

GS-W
& \underline{22.42} & \underline{0.792} & 0.379 
& \underline{23.66} & \underline{0.818} & 0.334 
& \underline{22.91} & \underline{0.750} & 0.406 
& \textbf{27.19} & \underline{0.898} & 0.238 \\

WildGaussians
& 20.82 & 0.724 & 0.393 
& 22.12 & 0.791 & \underline{0.289} 
& 19.90 & 0.695 & 0.375 
& 22.77 & 0.881 & \underline{0.184} \\

Ours(WilLaGS)
& \textbf{25.50} & \textbf{0.830} & \textbf{0.230}
& \textbf{27.07} & \textbf{0.879} & \textbf{0.200}
& \textbf{24.45} & \textbf{0.788} & \textbf{0.255}
& \underline{26.13} & \textbf{0.925} & \textbf{0.144} \\

\bottomrule
\end{tabular}
}
\end{table}

\textbf{Qualitative Results.} As shown in Fig.~\ref{fig:qual_pt} and \ref{fig:qual_osr}, WilLaGS consistently reconstructs finer geometric details and suppresses ghosting artifacts. On the PT dataset, baselines (e.g., K-Planes, GS-W) blur high-frequency structures, whereas WilLaGS recovers crisp \textit{Brandenburg Gate} pillars and \textit{Trevi Fountain} facade carvings. Similarly, on NeRF-OSR, our method successfully recovers intricate details that baselines entirely over-smooth, such as legible text (\textit{st}) on banners and thin tree branches (\textit{lwp}). Beyond geometric fidelity, our latent-conditioned field accurately models spatially-varying illumination (e.g., localized shadows and realistic lighting gradients in \textit{Sacre Coeur}). Conversely, prior methods collapse these into oversimplified global color shifts, yielding flat and unnatural appearances. This dual capability to restore crisp structures and plausible local lighting visually confirms our superiority.

\subsection{Ablation Studies}
To verify our core components, we conduct ablation studies alongside an overhead analysis. Compared to standard 3DGS, WilLaGS introduces only $\sim$933 MB of trainable VRAM ($\beta$-VAE: $\sim$ 66.64 MB, 3D field: $\sim$ 866.75 MB) and a static $\sim$528 MB for the frozen VGG-16 used by Teacher–Student perceptual mask.

\textbf{Effect of the VAE Latent Space.} To validate the $\beta$-VAE latent space, we replace it with discrete per-image embeddings as in NeRF-W \cite{martin2021nerfw}. As shown in Table \ref{ablation_table}, this substitution leads to a severe performance degradation across both datasets (e.g., a $\sim$4.36 dB drop in PSNR on PT and $\sim$2.82 dB on NeRF-OSR). This reveals that such independent deterministic embeddings merely overfit to individual images without learning a generalizable representation. In contrast, our $\beta$-VAE learns a probabilistic and shared manifold, enabling more robust and globally consistent appearance modeling.

\begin{figure}[t]
  \centerline{\includegraphics[width=\linewidth]{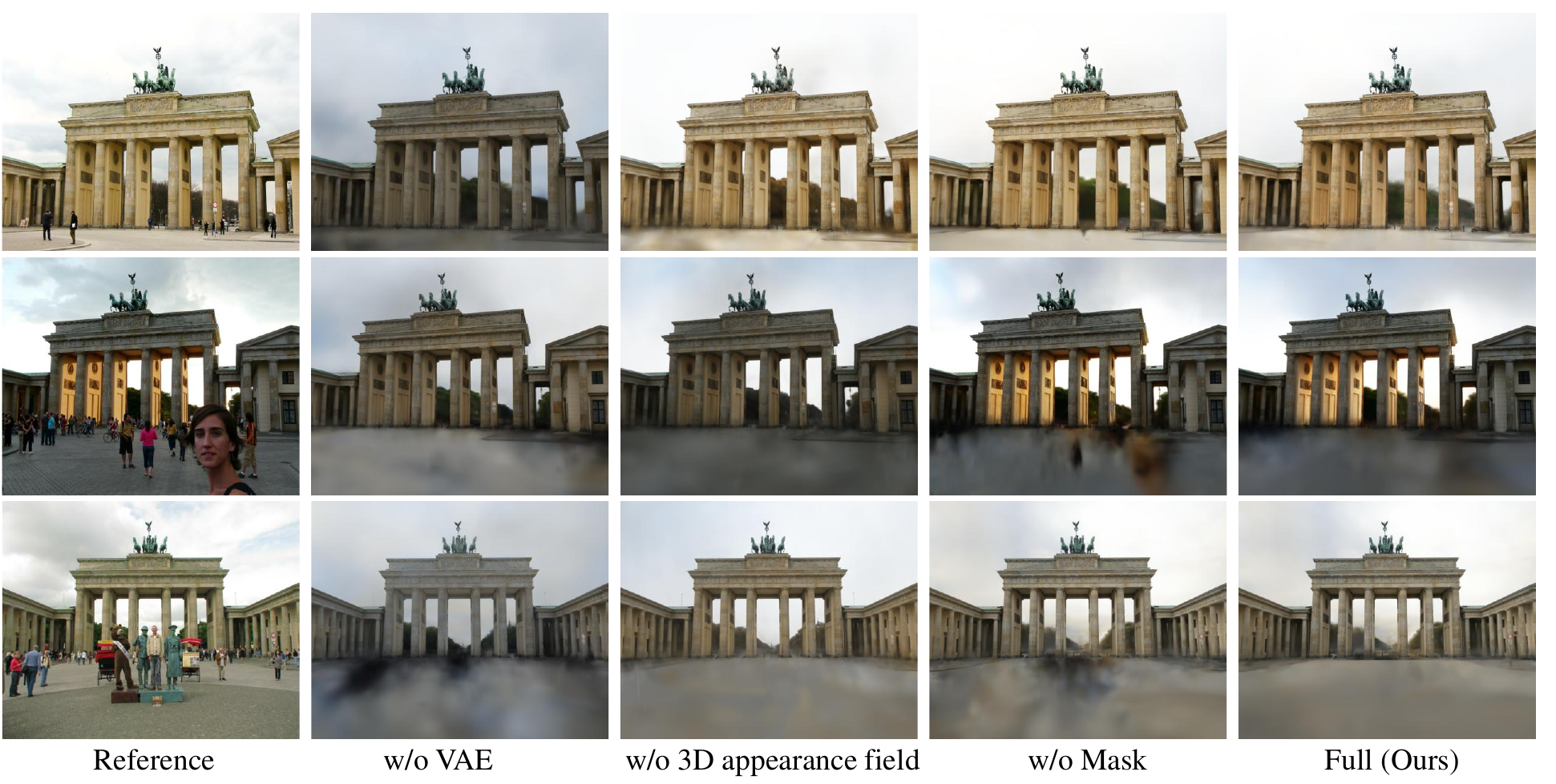}}
  \caption{\textbf{Qualitative ablation results}. Removing the VAE latent space leads to inconsistent color tones; omitting the 3D appearance field results in unnatural and spatially-uniform shadowing; disabling self-supervised masking causes severe ghosting artifacts.}
  \label{ablation}
\end{figure}

\begin{table}[t]
\centering
\caption{Ablation studies on key components. Metrics are averaged over all scenes in PT and NeRF-OSR datasets.}
\begin{tabular}{c|ccc|ccc}
\hline
& \multicolumn{3}{c|}{Photo Tourism} 
& \multicolumn{3}{c}{NeRF-OSR} \\ \cline{2-7}
& PSNR$\uparrow$ & SSIM$\uparrow$ & LPIPS$\downarrow$
& PSNR$\uparrow$ & SSIM$\uparrow$ & LPIPS$\downarrow$ \\ 
\hline
w/o VAE latent space     
& 21.15 & 0.845 & 0.177
& 22.97 & 0.822 & 0.213 \\

w/o 3D appearance field  
& 23.97 & 0.861 & 0.172
& 24.55 & 0.847 & 0.223 \\

w/o Teacher-student mask 
& \textbf{25.53} & 0.879 & 0.162
& 25.63 & 0.851 & 0.215 \\

Full (Ours)              
& 25.51 & \textbf{0.883} & \textbf{0.151}
& \textbf{25.79} & \textbf{0.856} & \textbf{0.207} \\ 
\hline
\end{tabular}
\label{ablation_table}
\end{table}

\textbf{Effect of the 3D Neural Appearance Field.} We evaluate the latent-conditioned 3D field by replacing it with a direct concatenation of the global latent code $z$ and intrinsic feature $f_i^s$.  Quantitatively, this causes noticeable drops in PSNR across both PT and NeRF-OSR (Table \ref{ablation_table}). Qualitatively, as depicted in Fig. \ref{ablation}, this spatially-uniform approach fails to capture localized illumination, producing unnatural and flat lighting instead of plausible shadows. 
 This confirms our Tri-Plane field is essential for translating global appearance codes into local and view-dependent 3D illumination.

\textbf{Effect of Teacher–Student perceptual masking.}  Finally, we disable our self-supervised masking module, forcing the model to fit all pixels, including transient elements. As clearly shown in Fig.~\ref{ablation} ("w/o Mask"), this leads to severe ghosting and blurry floaters from unmasked transient elements, corrupting the final static geometry. In contrast, our teacher-student EMA framework successfully derives a stable scene consensus to dynamically identify and suppress inconsistent regions, ensuring a cleaner static scene representation.

\begin{figure}[t]
  \centerline{\includegraphics[width=\linewidth]{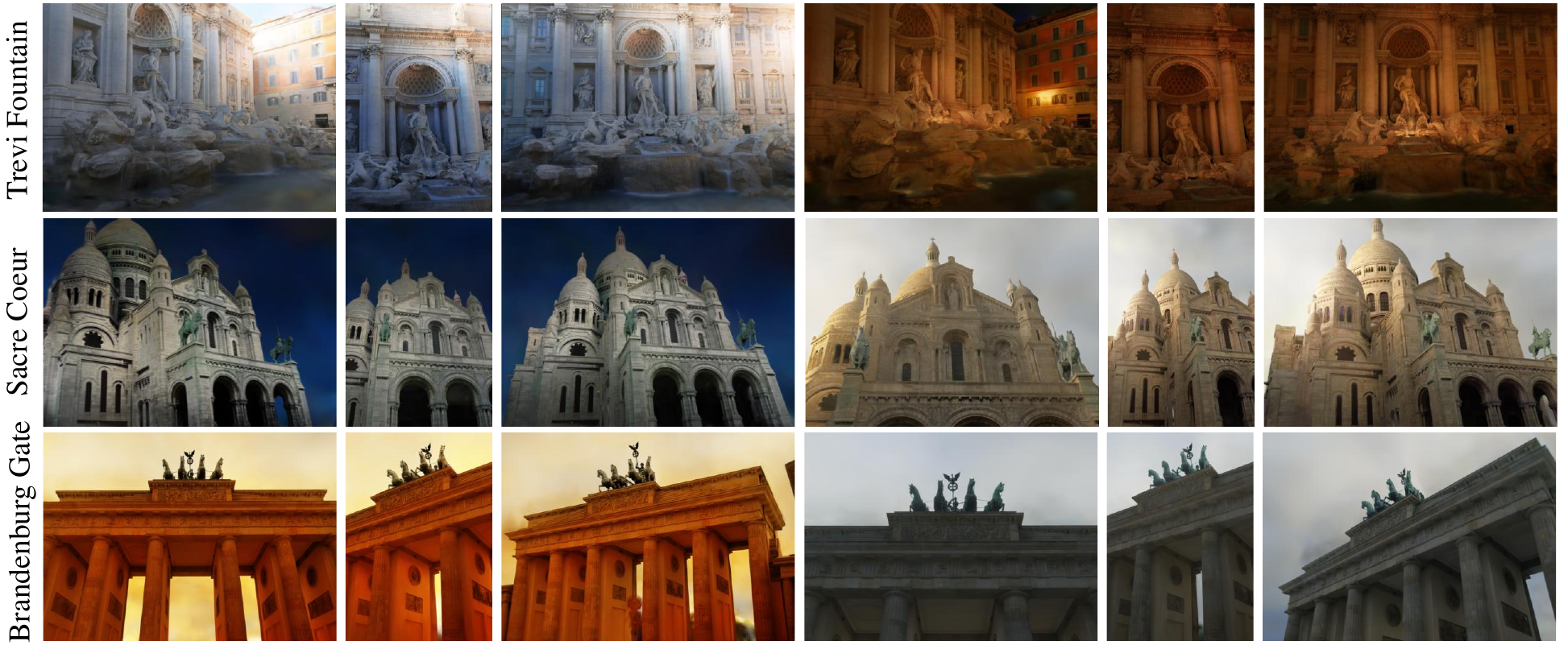}}
  \caption{\textbf{Multi-view appearance modeling.} Given a specific appearance latent code, WilLaGS synthesizes highly consistent novel views across different camera trajectories.}  
  \label{appearance}
\end{figure}

\subsection{Applications}
\subsubsection{Appearance Modeling and Transfer}
Our framework facilitates high-fidelity appearance modeling with strict multi-view consistency.  By conditioning the 3D appearance field on a specific latent code $z$, the scene can be re-rendered across arbitrary novel viewpoints under the corresponding lighting and weather conditions. As shown in Fig.~\ref{appearance}, WilLaGS produces novel views that maintain strong geometric and photometric consistency. For instance, sunlight correctly illuminates specific facades of the \textit{Trevi Fountain} while casting geometrically-consistent shadows across different camera poses, demonstrating our capability to produce faithful and spatially-varying illumination effects in 3D space.

\subsubsection{Appearance Interpolation}
\begin{figure}[t]
  \centerline{\includegraphics[width=\linewidth]{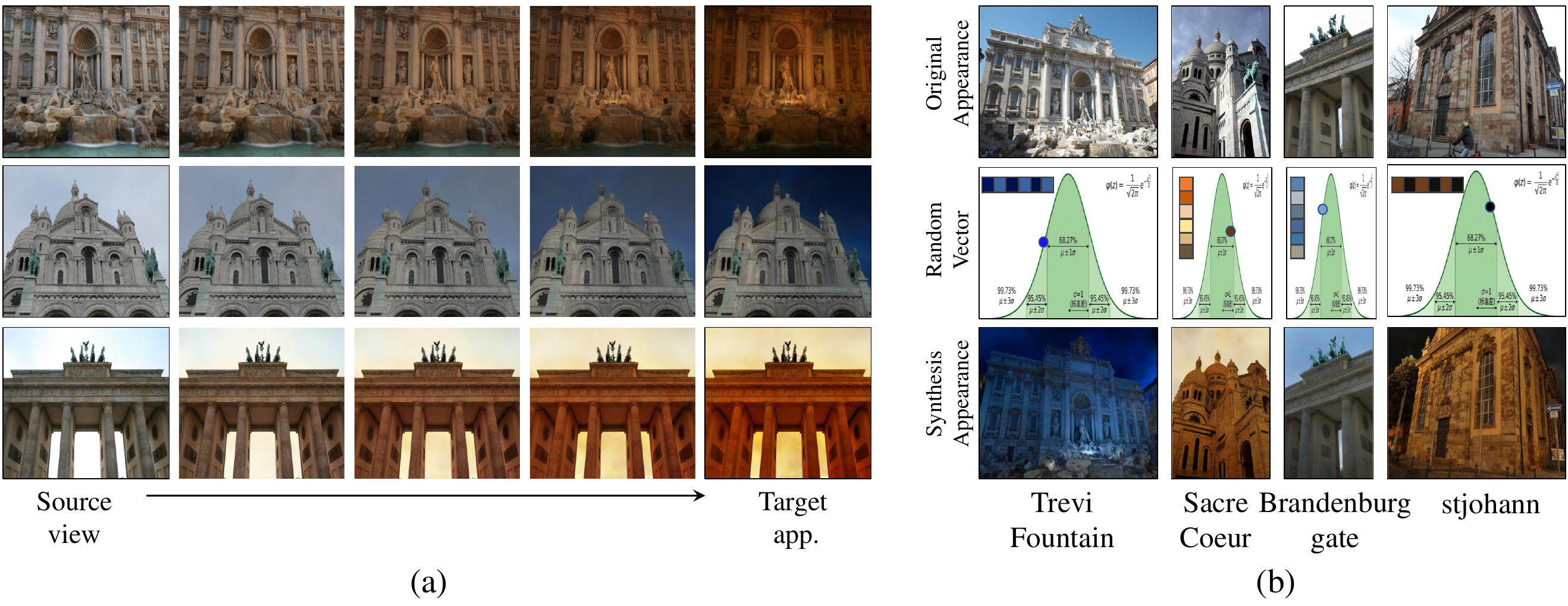}}
  \caption{(a) Appearance interpolation from a fixed viewpoint.(b) Novel appearance synthesis with random latent vectors from the learned standard Gaussian prior $\mathcal{N}(0, I)$. }  
  \label{app}
\end{figure}

WilLaGS has the capability to smoothly interpolate appearance between two latent codes $z_a$ and $z_b$ extracted from distinct reference images:
\begin{equation}
    z_{interp} = (1-\alpha)z_a + \alpha z_b,  \quad  \alpha \in [0, 1]
\end{equation}
As illustrated in Fig.~\ref{app}(a), feeding $z_{interp}$ into our conditional 3D appearance field generates realistic and consistent transitions between highly distinct conditions (e.g., sunny to night), while preserving the underlying scene structure.

\subsubsection{Unconditional Appearance Synthesis} Unlike prior methods relying on discrete embeddings or specific reference images for appearance transfer, our framework possesses inherent generative capabilities. By regularizing the latent space to follow a standard Gaussian prior, $p(z) = \mathcal{N}(0, I)$, our $\beta$-VAE enables unconditional appearance synthesis. As shown in Fig.~\ref{app}(b), for any independently trained scene, we can simply sample random vectors $z \sim p(z)$ and feed them into the latent-conditioned 3D appearance field. This process naturally generates diverse and physically plausible lighting and weather conditions (e.g., night, sunset, or overcast skies) without target image. This strongly demonstrates that WilLaGS learns a structured and true generative manifold, effectively disentangling complex appearances from scene geometry.

\section{Conclusion}
In this paper, we presented \textbf{WilLaGS}, a robust framework for in-the-wild 3DGS reconstruction.
Our approach introduces a generative latent-conditioned 3D neural appearance field, powered by a $\beta$-VAE, to model the complex and continuous manifold of scene appearance, together with a self-supervised teacher–student perceptual masking mechanism that dynamically suppresses inconsistent transient elements. 
Extensive experiments on PT and NeRF-OSR datasets demonstrate that \textbf{WilLaGS} achieves state-of-the-art reconstruction quality, significantly outperforming prior methods, while maintaining real-time rendering speeds. 
Furthermore, it enables diverse generative applications like appearance transfer and unconditional synthesis.  While handling dense occlusions and lacking explicit physical interpretability remain limitations, future work will explore semantic control over the latent manifold and expand to large-scale scenes.

\section*{Appendix: Evaluation Protocol}
In the original version of this paper, to ensure a consistent comparison environment, we utilized the default evaluation scripts provided by the GS-W \cite{zhang2024gsw} codebase to evaluate our WilLaGS and all other 3DGS-based baselines. We later identified that this default script performs full-image appearance optimization, which deviates from the strict left-half appearance optimization and right-half evaluation protocol established by NeRF-W \cite{martin2021nerfw}.

To ensure absolute rigor and fair comparison, the quantitative results reported in this updated arXiv version (including Table 1, Table 2, and the ablation studies) have all been rigorously re-evaluated under the strict NeRF-W half-image protocol. 

To provide full transparency to the community regarding the numerical impact of this protocol discrepancy, we report the comprehensive PSNR, SSIM, and LPIPS differences between the full-image and half-image evaluation protocols across all seven scenes for both GS-W and WilLaGS in Table \ref{tab:protocol_diff}. While the absolute metrics naturally differ (and in some specific scenes increase due to the inherent asymmetry of left/right image complexity), the relative performance trend remains robust, and WilLaGS continues to demonstrate state-of-the-art performance under both settings. We will explicitly release both evaluation scripts in our official codebase to ensure clear benchmarking for future researchers.

\begin{table}[h]
\centering
\caption{Comprehensive comparison between Full-image and strict Half-image evaluation protocols.}
\resizebox{\textwidth}{!}{
\begin{tabular}{@{} l @{\hspace{4pt}} c@{\hspace{4pt}}c@{\hspace{4pt}}c @{\hspace{8pt}} c@{\hspace{4pt}}c@{\hspace{4pt}}c @{\hspace{8pt}} c@{\hspace{4pt}}c@{\hspace{4pt}}c @{\hspace{8pt}} c@{\hspace{4pt}}c@{\hspace{4pt}}c @{}}
\toprule
 & \multicolumn{6}{c}{GS-W \cite{zhang2024gsw}} & \multicolumn{6}{c}{\textbf{WilLaGS (Ours)}} \\
\cmidrule(lr){2-7} \cmidrule(lr){8-13}
 & \multicolumn{3}{c}{Full-image} & \multicolumn{3}{c}{Half-image} & \multicolumn{3}{c}{Full-image} & \multicolumn{3}{c}{Half-image} \\
\cmidrule(lr){2-4} \cmidrule(lr){5-7} \cmidrule(lr){8-10} \cmidrule(lr){11-13}
Scene & PSNR$\uparrow$ & SSIM$\uparrow$ & LPIPS$\downarrow$ & PSNR$\uparrow$ & SSIM$\uparrow$ & LPIPS$\downarrow$ & PSNR$\uparrow$ & SSIM$\uparrow$ & LPIPS$\downarrow$ & PSNR$\uparrow$ & SSIM$\uparrow$ & LPIPS$\downarrow$ \\
\midrule
\multicolumn{13}{c}{\textit{Photo Tourism Dataset}} \\
\midrule
Sacre Coeur      & 22.23 & 0.836 & 0.291 & 19.78 & 0.834 & 0.261 & \textbf{25.84} & \textbf{0.891} & 0.164 & 23.80 & 0.889 & \textbf{0.150} \\
Trevi Fountain   & 22.93 & 0.773 & 0.217 & 20.11 & 0.763 & 0.237 & \textbf{24.54} & 0.789 & 0.207 & 24.27 & \textbf{0.822} & \textbf{0.168} \\
Brandenburg Gate & 26.86 & 0.921 & 0.243 & 23.86 & 0.891 & 0.183 & \textbf{29.94} & \textbf{0.939} & 0.139 & 28.45 & 0.937 & \textbf{0.134} \\
\midrule
\multicolumn{13}{c}{\textit{NeRF-OSR Dataset}} \\
\midrule
europa           & 23.31 & 0.833 & 0.335 & 22.42 & 0.792 & 0.379 & 24.38 & \textbf{0.843} & \textbf{0.216} & \textbf{25.50} & 0.830 & 0.230 \\
lwp              & 22.25 & 0.784 & 0.357 & 23.66 & 0.818 & 0.334 & 23.36 & 0.818 & 0.230 & \textbf{27.07} & \textbf{0.879} & \textbf{0.200} \\
st               & 23.13 & 0.754 & 0.398 & 22.91 & 0.750 & 0.406 & 23.43 & 0.763 & 0.267 & \textbf{24.45} & \textbf{0.788} & \textbf{0.255} \\
stjohann         & 25.72 & 0.894 & 0.276 & \textbf{27.39} & 0.898 & 0.238 & 25.55 & 0.906 & 0.161 & 26.13 & \textbf{0.925} & \textbf{0.144} \\
\bottomrule
\end{tabular}
}
\label{tab:protocol_diff}
\end{table}

\section*{Acknowledgements}

This work was supported by the Fundamental Research Funds for the Central Universities (No.~14380079), and the Collaborative Innovation Center of Novel Software Technology and Industrialization. 

%
%
\bibliographystyle{splncs04}
\bibliography{main}
\end{document}